\documentclass[10pt,twocolumn,letterpaper]{article}

\usepackage{cvpr}              

\usepackage{graphicx}
\usepackage{amsmath}
\usepackage{amssymb}
\usepackage{booktabs}
\usepackage{multirow}
\usepackage{slashbox}
\usepackage[cjk]{kotex}
\usepackage{xcolor}
\usepackage{amssymb}
\usepackage{pifont}
\usepackage{makecell}

\usepackage[pagebackref,breaklinks,colorlinks]{hyperref}

\usepackage[capitalize]{cleveref}
\crefname{section}{Sec.}{Secs.}
\Crefname{section}{Section}{Sections}
\Crefname{table}{Table}{Tables}
\crefname{table}{Tab.}{Tabs.}

\def\cvprPaperID{6298} 
\def\confName{CVPR}
\def\confYear{2023}

\begin{document}

\title{AVOID: The Adverse Visual Conditions Dataset with Obstacles \\ for Driving Scene Understanding}

\author{
Jongoh Jeong$^{2,*}$, 
Taek-Jin Song$^{1,*}$, 
Jong-Hwan Kim$^{1}$
Kuk-Jin Yoon$^{2}$,\\ 
    Robot Intelligence Technology Lab.$^{1}$, Visual Intelligence Lab.$^{2}$, KAIST\\
    {\tt\small tjsong@rit.kaist.ac.kr, \{jeong2, kjyoon\}@kaist.ac.kr, johkim@rit.kaist.ac.kr}
    \thanks{Taek-Jin Song and Jongoh Jeong contributed equally to this work.}
}

\maketitle

\begin{abstract}


Understanding road scenes for visual perception remains crucial for intelligent self-driving cars. In particular, it is desirable to detect unexpected small road hazards reliably in real-time, especially under varying adverse conditions (e.g., weather and daylight). However, existing road driving datasets provide large-scale images acquired in either normal or adverse scenarios only, and often do not contain the road obstacles captured in the same visual domain as for the other classes. To address this, we introduce a new dataset called AVOID, the \underline{A}dverse \underline{V}isual C\underline{o}nd\underline{i}tions \underline{D}ataset, for real-time obstacle detection collected in a simulated environment. AVOID consists of a large set of unexpected road obstacles located along each path captured under various weather and time conditions. Each image is coupled with the corresponding semantic and depth maps, raw and semantic LiDAR data, and waypoints, thereby supporting most visual perception tasks. We benchmark the results on high-performing real-time networks for the obstacle detection task, and also propose and conduct ablation studies using a comprehensive multi-task network for semantic segmentation, depth and waypoint prediction tasks~\footnote{Dataset will be released to public at [weblink].}. 

\end{abstract}

\section{Introduction}
\label{sec:intro}

\begin{figure}[!h]
    \centering
    \includegraphics[width=\linewidth]{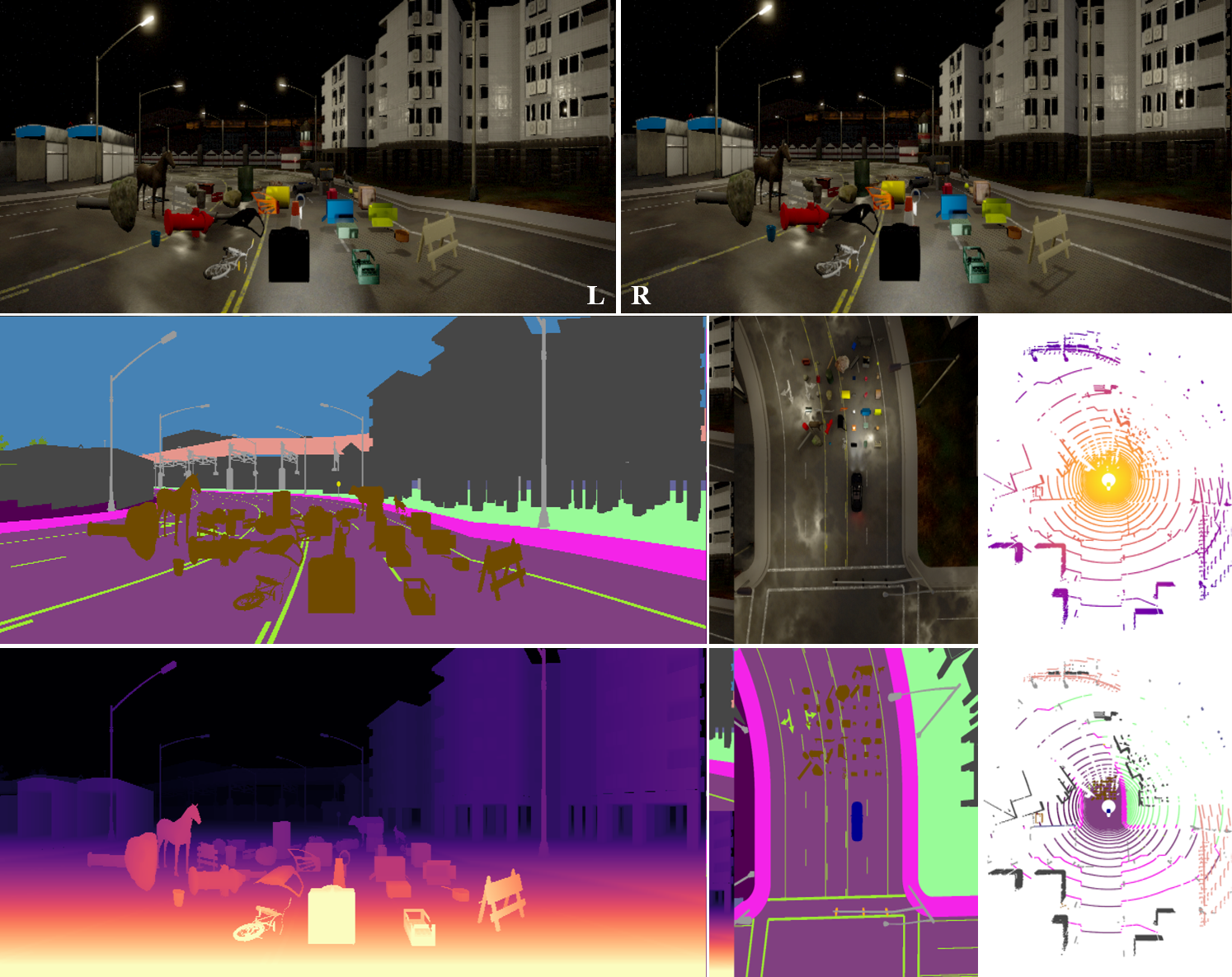}
    \caption{Sample views of all obstacles in AVOID. AVOID provides a set of multi-view images and annotations for each observed driving scene. As the vehicle navigates through the simulated routes, it collects a stereo pair of egocentric RGB images (\textit{top row}) and the corresponding annotations (semantic and depth maps, RGB and semantic maps in Bird's-Eye-View (BEV), and raw and semantic LiDAR data, shown from left to right on \textit{bottom two rows})
    }
    \label{fig:representative_fig}
\end{figure}

Visual understanding of urban scenes has been a growing field of research across academia as well as industry. In particular, the automation and its applications for fully autonomous vehicles have gained significant attention with a wide availability of urban street datasets with high-quality annotations for tasks such as semantic segmentation and depth estimation. More specifically, visually recognizing small, unexpected road hazards is a key issue in real-world driving scenarios in which unanticipated objects may interfere with the traffic at any given time and weather condition, leading to high collision risks. Moreover, the cases of failure to detect road debris on the highway have recently been on the rise, with about 1.3 million road traffic death tolls every year~\cite{WHO2021WHOFactSheet,Tefft2016AAATechReport, WHO2003WHOTechReport}, which the UN General Assembly has even set forth a plan to halve by 2030~\cite{USGA2020resolution}. Such road hazards are usually varied in size, shape, texture, and color, and tend to appear abruptly without clear hints in advance. Considering the safety of vulnerable road users including pedestrians, cyclists, and motorcyclists, these characteristics entail even greater significance in detecting unexpected road obstacles.

Objects in a driving scene, including obstacles, are encountered on the road not only during normal conditions, but also in ``unusual road or traffic conditions that [are] not known"~\cite{USMCSA05012020}, i.e., adverse driving conditions. Such conditions, including adverse weather (\textit{e.g.,} rain) and daytime (\textit{e.g.,} night), pose serious threats to road users and vehicles as they deteriorate desired lighting conditions for better visual perception and often incur driver fatigue~\cite{carey2017impact}. For instance, rainfall yields as high risk of crashes as 70\% compared to clear weather due to wet roads~\cite{andrey1993temporal}. Lighting, despite rarely being a direct cause of collisions, do compound more direct causal factors for driver fatigue by impeding the visual reaction time, thereby increasing stopping distance~\cite{carey2017impact, SULLIVAN2007638}. For self-driving applications, perceiving class-agnostic road hazards under these adverse conditions is thus of critical importance for the safety of both intelligent agents and practical road users.

To that end, we propose a novel simulated driving dataset containing on-road obstacles of various characteristics and the corresponding high-quality annotations from diverse sensor inputs. While there are a number of synthetic (simulated) and real-world datasets under adverse environmental settings only~\cite{SDV21acdc, Chitta2022PAMItransfuser, kerim2022semantic}, there is yet no single comprehensive obstacles dataset in such conditions that provides various output modalities including paired stereo images, semantic and depth maps, and raw and semantic LiDAR sensor data. Several previous works~\cite{romera2017erfnet, song2022rodsnet}, on the other hand, address the task of detecting small unexpected on-road obstacles by opting to fuse two different segmentation datasets in similar real-world road driving domains, but only in clear weather. We thus introduce a novel obstacles dataset under various adversarial scenarios in a simulated environment for use in most visual perception tasks including semantic segmentation.  Note that the obstacles in our dataset as seen in Fig.~\ref{fig:representative_fig} are ones encountered commonly on the road and we exploit the CAR Learning to Act (CARLA) simulator~\cite{Dosovitskiy17CARLA} to realize various scenes that are otherwise not feasible for data collection in real worlds due to physical resource and societal costs.


We highlight our contributions in three-fold as follows:
\vspace{-0.1in}
\begin{enumerate} \itemsep -4pt
    \item We introduce a novel large-scale dataset for obstacle detection under adverse environmental conditions, collected in a simulated environment. Our dataset provides high-quality annotations for use in most visual perception tasks that require RGB, Depth and LiDAR as inputs. 

    \item We provide benchmark results on the obstacle detection task under 42 different adverse conditions on high-performing single-task networks for the semantic segmentation and waypoint prediction tasks.
    
    \item We examine the performance of a real-time, lightweight obstacle detection network on our proposed dataset for road navigation using a transformer-based feature fusion approach, achieving up to 51.81\% IoU in obstacle detection (ResNet-18 backbone) at 18.87 FPS on a single Nvidia RTX 3090 GPU.
    
\end{enumerate}
\section{Related Work}
\label{sec:relatedwork}

\noindent \textbf{Unexpected small on-road obstacle detection.} \quad
%
%
The task of detecting unexpected on-road obstacles deals with identifying never-seen-before objects from the semantic segmentation map in a driving scene. As it is a time- and safety-critical task that requires efficient, real-time processing in practice, a number of proposed vision-based comprehensive systems~\cite{rateke2022passive, ismail2013monocular, ma2009real} have thus incorporated various image views, such as passive vision, frontal and rear-view images and shadows of objects on the road. Similar to several other obstacle detection works in light of anomaly detection~\cite{lis2019detecting, ohgushi2020road, lakshminarayanan2017simple, creusot2015real} based on resynthesis, a restricted Boltzmann machine and an uncertainty-based model, these systems, however, merely target detecting the presence of obstacles in the free space in front or at the back of the vehicle, in addition to detecting other vehicles and pedestrians. 

It is, yet, of paramount importance to identify and localize \textit{unexpected} on-road obstacles of various characteristics (\textit{e.g.,} size, shape, and texture) at pixel-level from the segmentation map, in particular, those at a long distance range for timely response to potential road hazards\cite{pinggera2016lostandfound, pinggera2015high}. For safety-critical moving platforms like self-driving vehicles, this task thus remains a key component that can further minimize traffic collisions due to undesirable road debris and obstacles. To that end, traditional filtering-~\cite{godha2017road} and recent deep learning-based~\cite{Salavati2018googlenet, lis2022perspective} approaches have been proposed to distinguish relevant objects from the irrelevant ones in a driving scene using image inputs. 

Using a Light Detection and Ranging (LiDAR) sensor that has recently gained attention as a standard sensor for various autonomous driving tasks, \cite{levi2015stixelnet} introduces the Stixel representation of vertical strips in the RGB image and \cite{tan2022road} dynamically selects the threshold parameters for an improved DBSCAN clustering for obstacle detection. However, LiDAR sensors incur significant physical maintenance and computational overheads, and thus many other works have instead opted to use cost-effective RGB and depth sensors together in cross-modal learning~\cite{costa2012obstacle, fan2019multiple}. In particular, \cite{sun2020real, ramos2017detecting} demonstrate the advantage of contextual learning across data modalities using semantic and geometry cues for obstacle detection from the segmentation map. Further, \cite{song2022rodsnet} achieves higher detection performance without the depth input by refining the predicted semantic and disparity maps in a single stage.



\noindent \textbf{Driving datasets in adverse conditions.} \quad
Apart from employing different sensors for more comprehensive understanding of a scene in an adverse environment, the quantity and quality of vision datasets for self-driving primarily determine the predictive performance of an intelligent driving agent. For the task of semantic segmentation, a task of classifying each pixel in an image into its representative category among the pre-defined set of classes, KITTI~\cite{geiger2012we} and Cityscapes ~\cite{Cordts2016cityscapes} datasets have sparked significant progress in learning-based visual recognition in the real domain given RGB camera and LiDAR sensor inputs. Following the two seminal ones, a number of other datasets aimed at increasing in scale have demonstrated the difficulty of acquiring high-quality pixel-level annotations by hand~\cite{huang2019apolloscape, neuhold2017mapillary}, and accordingly, synthetic datasets of even larger scales~\cite{johnson2016driving, richter2017playing, richter2016playing} have shown to be advantageous in collecting and annotating the ground truth 
data automatically. 

Several simulator-based~\cite{deschaud2021pariscarla3d, Chitta2022PAMItransfuser} explicitly exploit the use of the CARLA simulator~\cite{Dosovitskiy17CARLA} on the Unreal Engine for simulating visual road navigation in an adverse environment set by the pre-defined environmental parameters. Other works synthetically generate adverse conditions by applying the fog density by adjusting the control parameter, $\beta$~\cite{Sakaridis18cityscapesfoggy}, and use a generative model to adapt from clear to adverse weather conditions~\cite{rothmeier2021let}. However, most datasets including Foggy Driving~\cite{Sakaridis18cityscapesfoggy}, Foggy Zurich~\cite{dai2020curriculum}, Nighttime Driving~\cite{dai2018dark}, Dark Zurich~\cite{sakaridis2020map}, 
 Raincouver~\cite{tung2017raincouver}, WildDash~\cite{zendel2018wilddash} and BDD100K~\cite{yu2020bdd100k} aimed at semantic labelling in adverse conditions (\textit{e.g.,} fog, night, rain and snow) merely contain fewer than a thousand images, with \cite{yu2020bdd100k} being the only exception with 1,346 images but with severe errors. A more recent \cite{SDV21acdc} contains an equally distributed set of real-world adverse condition images following the same four adverse classes at a larger scale, and \cite{Chitta2022PAMItransfuser} is only trained with coarse semantic labels with the primary task being road navigation. To address the limitations of existing adverse conditions datasets from the obstacle detection perspective~\cite{kenk2020dawn, musat2021multi, bijelic2020seeing, pfeuffer2019robust, sheeny2021radiate}, our dataset provides a more diverse and large-scale sensor input data acquired in the same visual domain, and the corresponding pixel-level class-agnostic obstacle annotation, together with the other benchmark classes.

\section{AVOID Dataset}
\label{sec:method}
\begin{figure}[!t]
    \centering
    \includegraphics[height=8cm, width=\linewidth]{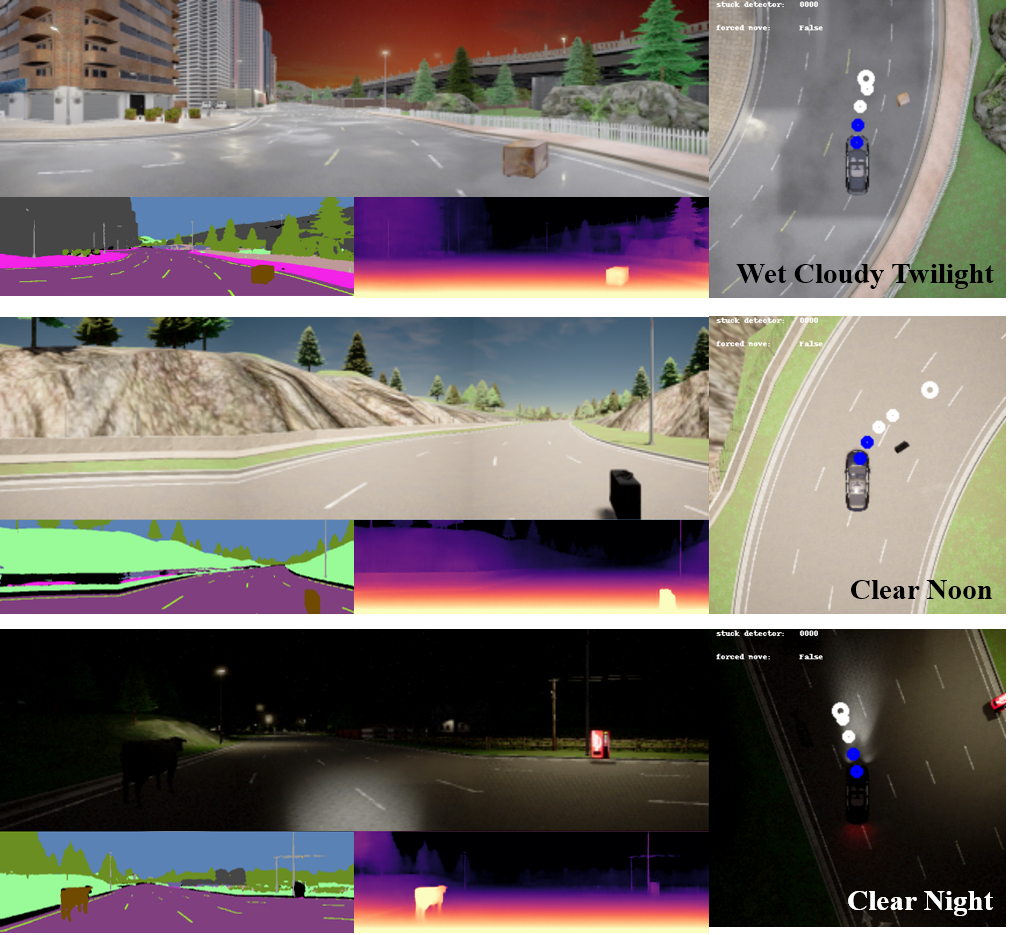}
    \caption{Samples results of the vehicle avoiding obstacles ahead (a nightstand, a black suitcase, and a black cow, from \textit{top} to \textit{bottom} in order).}
    \label{fig:sample_obstacle_avoidance}
\end{figure}

\begin{table*}[!t]
    \centering
    \caption{Comparison of AVOID against clear and adverse environmental condition semantic segmentation datasets}
    \resizebox{\linewidth}{!}{%
    \begin{tabular}{ccccccccccc}
    \toprule
    
    \multirow{2}{*}{{\bf Dataset}} & \multirow{2}{*}{{\bf Env.}} & \multirow{2}{*}{{\bf Condition}} & \multirow{2}{*}{{\bf Modality}} & \multicolumn{2}{c}{{\bf Semantic Annotation}} & \multicolumn{2}{c}{\bf Depth Map} & \multirow{2}{*}{{\bf Waypoints}} & \multirow{2}{*}{{\bf Resolution}} & {\bf \# Frames}\\ 
    & & & & \footnotesize{\# Classes} & \footnotesize{Obstacle} & Acquisition & Precision & & & Train / Val. / Test\\
    \midrule
    
    Cityscapes~\cite{Cordts2016cityscapes} & \multirow{3}{*}{\rotatebox[origin=c]{90}{\footnotesize{Real}}} & Clear & Stereo RGB & Fine (19) & {\color{red}{\ding{55}}} & SGM~\cite{hirschmuller2005accurate} & - & {\color{red}{\ding{55}}} & 2048$\times$1024 & 5,000 (2,975 / 500 / 1,525)\\
    
    Lost and Found~\cite{pinggera2016lostandfound} & & Clear & Stereo RGB & Coarse (3) & \ding{51} & SGM~\cite{hirschmuller2005accurate} & - & \color{red}{\ding{55}} & 2048$\times$1024 & 2,239 (1036 (train/val.) / 1203) \\
    
    ACDC~\cite{SDV21acdc} &  & Adverse & Monocular RGB & Fine (19) & {\color{red}{\ding{55}}} & \multicolumn{2}{c}{{\color{gray}{\footnotesize{N/A}}}} & \color{red}{\ding{55}} & 1920$\times$1080 & 4,006 (1,600 / 406 / 2,000)\\

    \midrule 

    TransFuser~\cite{Chitta2022PAMItransfuser} & \multirow{3}{*}{\rotatebox[origin=c]{90}{\footnotesize{Sim.}}} & Adverse & Monocular RGB & Coarse (7) & {\color{red}{\ding{55}}} & Depth camera & FP24~\cite{Dosovitskiy17CARLA} & \ding{51} & 960$\times$160 & 226,889 (168,159 / 58,730 / -)\\ 

    SHIFT~\cite{shift2022Sun} & & Adverse & Stereo RGB & Fine (23) & {\color{red}{\ding{55}}} & Depth camera & FP24~\cite{Dosovitskiy17CARLA} & {\color{red}{\ding{51}}} & 1280$\times$800 & 2.5 M \\        
    
    AVOID (Ours) &  & Adverse & Stereo RGB & Fine (17) & \ding{51} & Depth camera & FP24~\cite{Dosovitskiy17CARLA} & \ding{51} & 960$\times$320 & 21,403 (17,792 / 2,477 / 1,134) \\  
    
    
    \bottomrule
    \end{tabular}
    }
    \label{tab:dataset_comparison}
\end{table*}

\begin{figure*}[!t]
    \centering
    \includegraphics[width=\linewidth]{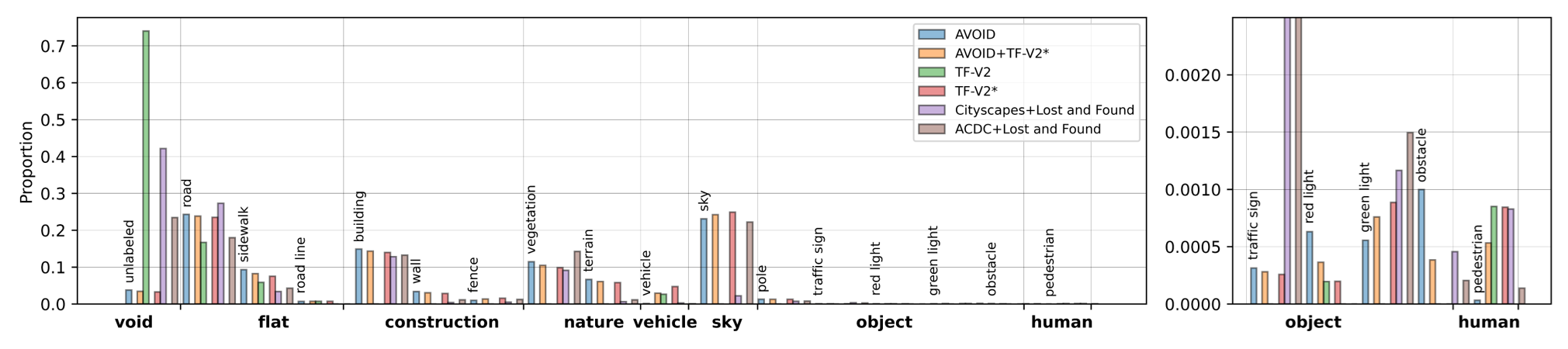}
    \caption{Comparison of data distributions for pixel-wise semantic labels across the segmentation datasets.}
    \label{fig:data_distribution_comparison}
\end{figure*}

We follow the design principles, and the data collection and annotation strategies established by the well-received semantic segmentation datasets~\cite{Cordts2016cityscapes, SDV21acdc} to target our primary task, obstacle detection, as well as by the seminal simulator-based datasets for road navigation~\cite{Dosovitskiy17CARLA, deschaud2021pariscarla3d, Chitta2022PAMItransfuser} to realize and collect more practical data under adverse condition scenarios. 

\noindent 
\textbf{Motivation.} \quad
%
With the growing demand for driving scene datasets that target domain shifts from synthetic and real datasets, driving scene datasets under adverse conditions are becoming essential. However, only a few such datasets containing road obstacles are available in real-world settings due to practical concerns in data collection, let alone in simulated environments. 
In light of the success of recent CARLA simulator-based road navigation datasets under adverse conditions~\cite{Chitta2022PAMItransfuser, deschaud2021pariscarla3d} that provide multiple output modalities including LiDAR and RGB, we thus build upon their works to introduce class-agnostic obstacles of various sizes, shapes, textures, and categories to the provided town routes in CARLA. In common visual perception tasks for road navigation, unknown obstacles are bound to appear in unanticipated time, place and occasions without hints, thereby restricting intelligent driving agents in both navigating planned routes and safely avoiding unexpected objects appearing in their fields-of-view. As fully autonomous vehicles are becoming rapidly commercialized, detecting 
unexpected obstacles as small as 5 cm in height~\cite{romera2017erfnet, song2022rodsnet} is a key specification that needs to be met in practical road driving scenarios.

\noindent \textbf{Approach.} \quad
A common approach to collecting an urban scene driving dataset with road obstacles is by simulating a driving environment virtually as considerable costs in physical resources, time, personnel and safety are inevitably associated with collecting real-world road scenes containing diverse on-road obstacles. The Town05 CARLA dataset~\cite{Chitta2022PAMItransfuser} is one that is collected through simulated navigation and evaluated by taking simulated RGB images and LiDAR data as inputs to their multi-modal fusion transformer network to encourage global contextual learning across different data modalities. 

While their data collection strategies include variations in weather and daylight conditions divided into 36 different adversarial conditions following~\cite{Chitta202NEATICCV}, and using multi-modality sensors to precisely describe the scene, Chitta \textit{et al.} overlook the unrealistic settings in the collected scenarios in which the environmental condition is altered at every frame during the data collection process. Moreover, \textit{clear} weather is omitted to evaluate specifically under the select CARLA's adversarial scenarios 1,3,4,7,8,9,10 generated according to the NHTSA pre-crash typology, such as lane merging and changing, as well as negotiations at traffic intersections and roundabouts, as per the CARLA Autonomous Driving Challenge scenarios~\footnote{\href{https://leaderboard.carla.org/scenarios}{https://leaderboard.carla.org/scenarios}}.

\begin{figure}[!t]
        \centering
        \includegraphics[width=\linewidth]{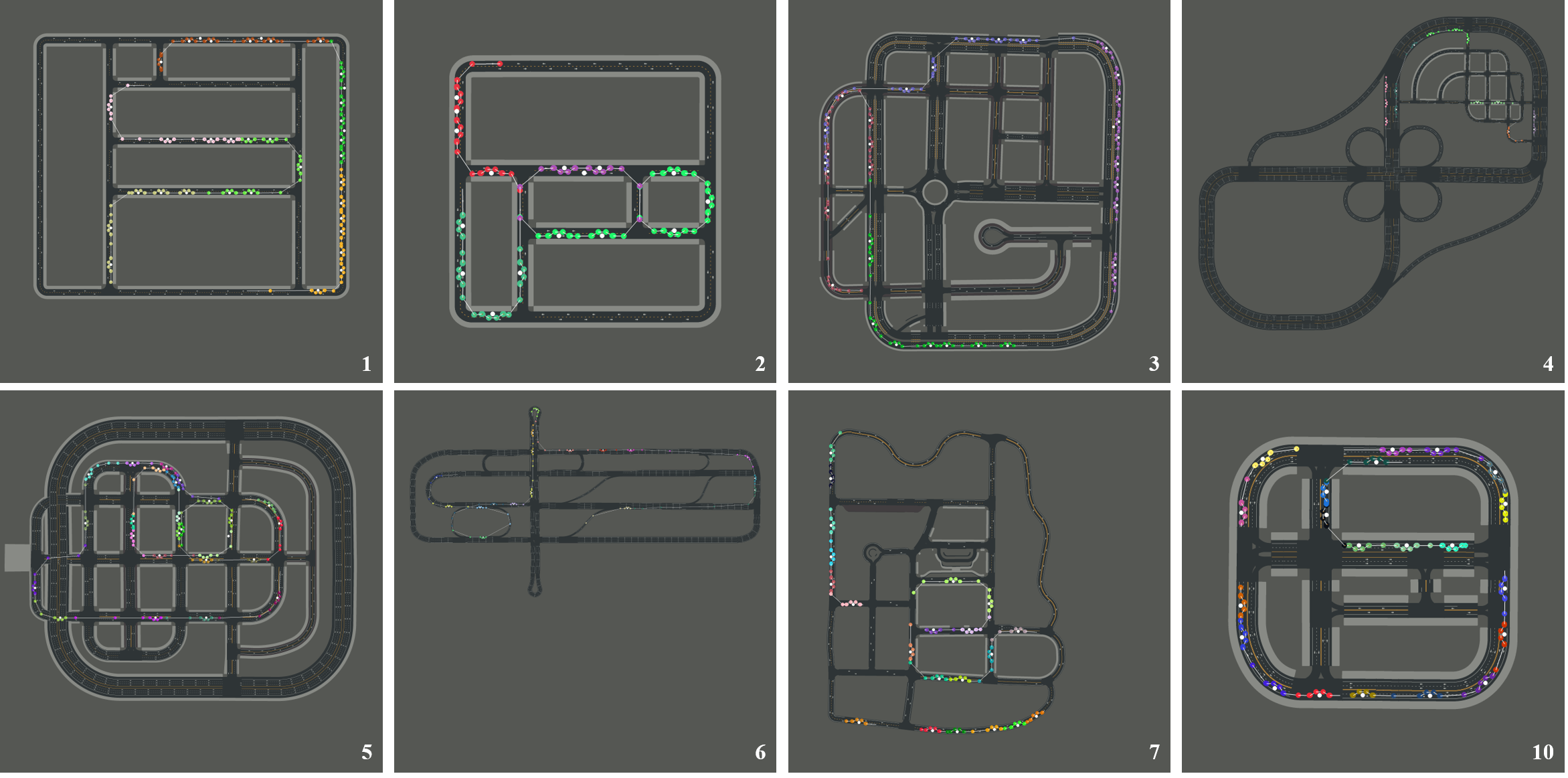}
        \caption{Town routes (Train: 1--4, Val.: 5, 6, Test: 7, 10) in the CARLA Simulator. Note that only one sample from each town is shown.}
        \label{fig:towns}
    \end{figure}

We thus introduce a new dataset called AVOID, the \underline{A}dverse \underline{V}isual C\underline{o}nd\underline{i}tions \underline{D}ataset, that builds upon the previous CARLA-based one~\cite{Chitta2022PAMItransfuser} for better scene perception by providing a more comprehensive set of annotations that are acquired with time synchronization (\textit{e.g.,} paired stereo images, semantic LiDAR data, and additional pixel-level semantic class labels). Building upon the code~\cite{chen2020learnbycheat} based on the handcrafted rules and guided by the collection principles by the authors of \cite{Chitta2022PAMItransfuser}, we carried out an expert policy provided by the CARLA simulator which allows the vehicle to navigate with the privileged ground truth waypoint labels. These waypoints set by the simulator are then navigated through the Towns as shown in Fig.~\ref{fig:towns} using an A* planner followed by two independent PID controllers in the lateral and longitudinal directions. The high-quality pixel-level annotations for the sensor outputs, including paired stereo RGB images, semantic and depth maps, and raw and semantic LiDAR data, have been auto-labeled by the simulator.
We highlight the improvements as follows: (1) addition of \textit{clear} weather condition, yielding 42 different weather-daytime combinations, (2) fixed adversarial environment for each scenario for more realistic environmental changes, rather than altering on a frame-by-frame basis, (3) inclusion of the \textit{obstacle} semantic class, (4) synchronized stereo RGB image pairs in a panoramic view (stitched \textit{Left}, \textit{Focus} (Center), \textit{Right} views with each camera FoV=60$^{\circ}$ and baseline distance 0.2m apart), and (5) more diverse data modalities including paired RGB and semantic LiDAR acquired synchronously. Detailed comparison across related datasets and annotation statistics are presented in Table~\ref{tab:dataset_comparison} and Fig.~\ref{fig:data_distribution_comparison}, respectively. 

\noindent 
\textbf{Obstacles.} \quad
We acquired 45 different 3-D object models of various textures, shapes and sizes in FBX format from the web~\footnote{FBX (.fbx) is a file format for high-fidelity 2D and 3D geometry and animation data, widely used in film, game, and AR / VR development.}, including those belonging to categories ranging from \textit{Nature}, \textit{Construction}, \textit{Transport Artifact}, \textit{Container} and  \textit{Animal} to \textit{Other} (See \textit{Supplementary} for detailed description of objects). We specifically selected objects that are encountered commonly on-the-road, roadside, and random items that may fall from cargo trucks from behind.

For each imported object, we manually applied transforms (\textit{e.g.,} translation, rotation, uniform scaling) to fit to the road environment. For example, a fire hydrant object was uniformly scaled by 1.8, translated by (0,0,25) and rotated by (0,90,-60) tuning parameters. The object positions on the path were set according to (1) whether the vehicle travel path is heavily interrupted due to the object at the intersection, for instance, (2) whether the vehicle can safely follow the waypoints around the object, and (3) if the vehicle encounters the object at some distance into straight or curved roads. We accordingly set the waypoints for the vehicle to avoid obstacles as follows: (1) using the obstacle detector in CARLA, vehicle is notified of the obstacle at 10 meters in distance, (2) waypoints to circumvent the obstacle are manually set such that they are 3 meters away from the obstacle, and (3) straight waypoints in the forward direction after avoiding the obstacle are set after the vehicle is 5 meters apart from the passed obstacle. The type of obstacle for each path in a scenario is randomly determined for more realistic considerations as well as convenience. We note that this simulation environment can be further expanded by customizing with objects of the user's interest. We visualized in Fig.~\ref{fig:sample_obstacle_avoidance} sample road navigation views in which the ego-centric vehicle avoids the obstacle in its frontal view by circumventing it.

\subsection{Data Specifications}
    

    Our data collection process is automated in CARLA~\cite{Dosovitskiy17CARLA} 0.9.11 simulator which provides eight publicly available town maps.
    We take advantage of the CARLA environment for its customize-able weather and temporal parameters in simulated self-driving scenarios. We further detail these adverse condition parameters in \textit{Supplementary}.

    Throughout the path routes, our data collection process was synchronized in time for all resulting RGB and LiDAR data. In the simulated environment, we set the LiDAR sensor parameters \{Dropoff General Rate=0, Dropoff Intensity=10, Dropoff Zero Intensity=0\} in order to match the raw and semantic LiDAR data for each frame. For the RGB camera, the baseline distance, field-of-view, and the width are set to 0.2 meters, 60$^{\circ}$, 320 pixels, respectively. The maximum, visible depth range in an acquired image is thus 55.4 meters, according to focal length $f = (width/2) / tan(FoV/2)$ and the epipolar geometry equation for depth and disparity. 

    For safe navigation by avoiding obstacles, we intended to allow the vehicle to cross the center yellow line upon encountering an obstacle by manually setting waypoints accordingly. In this approach, we limited the appearance of pedestrians and other vehicles in the dataset in order for the vehicle not to confuse with or stop abruptly in front of people and vehicles. We note that this pre-defined setting resulted in relatively lower numbers of the two classes compared to those in \cite{Chitta2022PAMItransfuser}. Moreover, the image height is experimentally determined to be twice that of the TransFuser as the field-of-view is significantly restricted, or cut out, for our panoramic view of 960$\times$320 resolution when a pedestrian of short height or small obstacles appear. During navigation, the vehicle is also set to stop and remain idle if it detects a red traffic light within its driving angle of 60$^{\circ}$ or 7 meters in distance. 
    
    
    

\noindent
\textbf{Adverse Environmental Conditions.} \quad
AVOID involves 7 different weather (Clear, Cloudy, Wet, Wet-Cloudy, Hard-rain,  Mid-rain, Soft-rain) and 6 different daytime conditions (Night, Twilight, Dawn, Sunset, Morning, Noon), totaling 42 unique adverse conditions that are realized per scenario. The specific parameters tuned for each condition are outlined in \textit{Supplementary}, while the remaining fog density, distance, falloff, wetness parameters are set to zero.

\noindent
\textbf{Dataset Split.} \quad
%
AVOID is split into three subsets; the training and validation subsets respectively are derived from the train and validation subsets of the Longest6 benchmark~\cite{Chitta2022PAMItransfuser}, with obstacles appearing in the driving scenes overlapping in train and validation subsets. The test subset contains paths from the Towns 7 and 10 from the eight CARLA towns, containing obstacles that are not part of either train or validation. 
The Longest6 benchmark contains 1.5 km long routes, higher traffic density, and challenging traffic scenarios set by the authors, from which we selected a smaller portion among the available towns 1-6 excluding areas of infeasible waypoint routes by the vehicle upon observing an obstacle. 
The waypoints for navigation in the test scenarios were all hand-labeled to specifically design challenging paths containing the obstacles that do not appear in the other two subsets. This allows for evaluating the capability of a navigation model to detect even untrained obstacles. In each of the train-val/test subsets, we introduce 39/6 unique obstacles which are described in detail in \textit{Supplementary}.







\section{Single-task Benchmark Evaluation}
\label{sec:experiments}

\begin{figure*}[!ht]
    \centering
    \includegraphics[width=\linewidth]{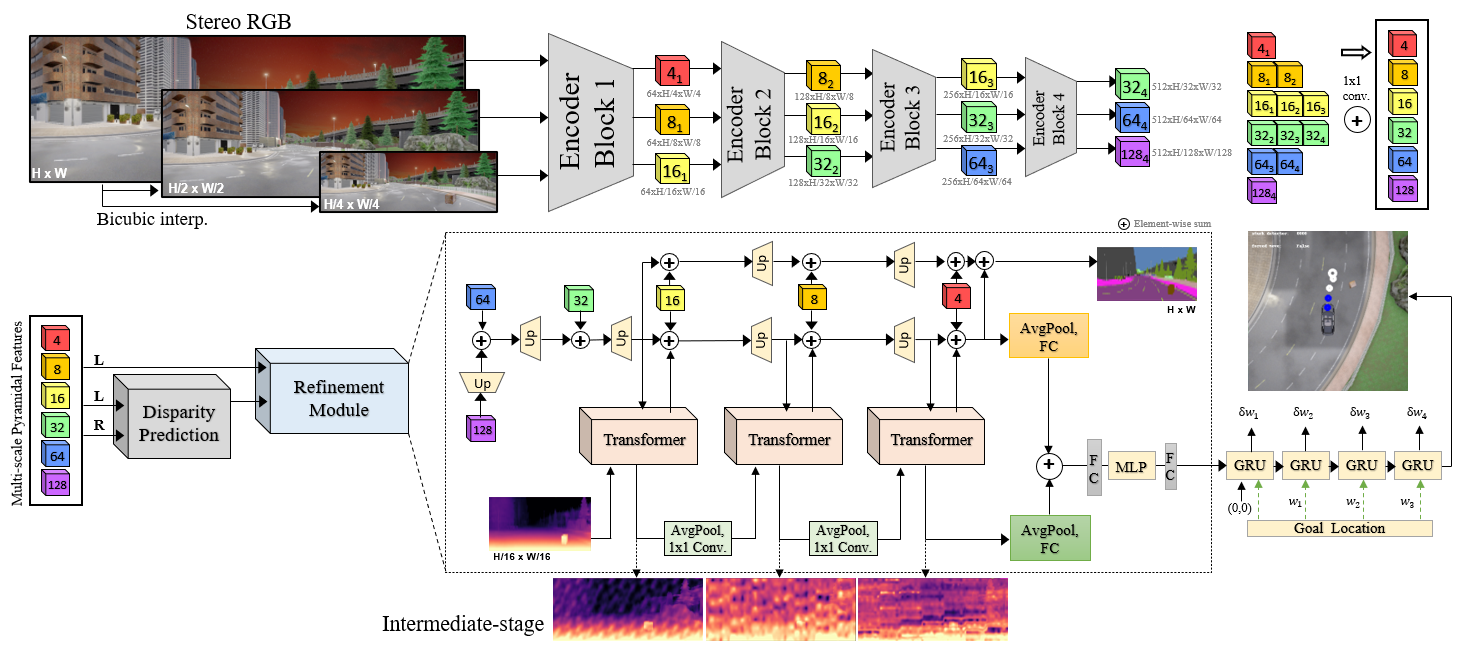}
    \caption{Overall network architecture of our multi-modal (RGB-D) transformer-based fusion network for multi-task learning, called SwiftFuser. We compare this network to RODSNet with GRU added for waypoint prediction, in which initial semantic and disparity maps are each averaged pooled and input to the GRU modules.}
    \label{fig:overview}
\end{figure*}
    
\subsection{Supervised Semantic Segmentation on AVOID}
    
    As AVOID is designed to incorporate obstacles of various characteristics for detection from the segmentation map, it supports the standard semantic segmentation task. All benchmark results on high-performing semantic segmentation methods evaluated on the test set of AVOID are reported in Table~\ref{tab:real_time_semantic_segmentation}, and the detailed per-class results for the state-of-the-art obstacle detection networks are reported in Table~\ref{tab:semseg_results}. While the current state-of-the-art obstacle detection network, RODSNet, achieves the highest overall mIoU across all 17 tested classes, RODSNet with a waypoint prediction network (GRU) outperforms for the obstacle class. Our experimented SwiftFuser model yields the second-best score for the obstacle class, while achieving a comparable overall mIoU. 
    We highlight that the segmentation results for \textit{pedestrian} and \textit{traffic light for stopping} are close to zero due to few occurrences of such class objects encountered before the vehicle throughout the test dataset.
    
    \noindent
    \textbf{Evaluation Metric.} \quad
    We evaluate the semantic segmentation performance quantitatively by the Intersection-over-Union (IoU) metric. IoU, or commonly the Jaccard index, accounts for the ratio of the area of overlap to union between the predicted and the ground truth image pixels as described in Eq.~\ref{eq:iou}.
    \begin{equation}
        IoU = \frac{TP}{TP+FP+FN},
        \label{eq:iou}
    \end{equation}
    \noindent where TP, FP, FN, respectively, denote true positive, false positive, and false negative pixels. The class-averaged IoU is represented as the mean IoU, or mIoU.
    
    \begin{table*}[!t]
        \centering
        \caption{Per-class semantic segmentation IoU (\%) results on test set of the AVOID dataset, averaged over all conditions. Class labels: Building, Fence, Pedestrian, Pole, Road Line, Road, Sidewalk, Vegetation, Vehicle, Wall, Sky, Terrain, Traffic Light for Stopping (Red/Yellow Light), Obstacle, Traffic Light for Proceeding (Traffic Light/Green Light), and Traffic Sign (from left to right). Best in \textbf{boldface}, runner-up in \underline{underline}. See \textit{Supplementary} for condition-specific results.}
        \resizebox{2.05\columnwidth}{!}{
        \begin{tabular}{lcccccccccccccccc|c}
        \toprule
        \textbf{Method} & Bui & Fen & Ped & Pol & RoLi & Roa & Sid & Veg & Veh & Wal & Sky & Ter & TL-s & {\bf Obs} & TL-p & TS & \textbf{mIoU} \\
        \midrule

        RFNet~\cite{sun2020real_rfnet}  & 43.63 & \textbf{19.21} & 0.00 & \textbf{42.74} & 13.47 & 88.67 & 85.51 & 63.48 & 0.00 & 0.00 & 0.00 & \textbf{56.17} & 0.00 & 23.61 & 0.00 & 0.00 & 33.50          \\
        
        SwiftNet$^{\dagger}$~\cite{orvsic2021efficient}  & \underline{80.95
        }& 3.93 & 0.00 & 27.94 & 54.22 & 89.40 & 68.62 & \textbf{69.47} & 8.59 & 1.93 & 92.67 & 43.38 & 0.00 & \underline{42.47} & 16.11 & 6.79 & 36.59  \\
        
        RODSNet$^{\dagger}$  & 78.15 & 2.71 & 0.00 & \underline{30.65} & \underline{60.80} & 85.65 & 65.36 & 69.30 & 12.71 & \textbf{9.98} & 91.80 & 42.31 & \textbf{1.05} & 40.47 & \underline{19.90} & \underline{13.12} & 37.62           \\
        RODSNet$^{\ddagger}$  & \textbf{81.50} & 5.16 & 0.00  & 29.97 & 56.95 & \textbf{91.52} & \textbf{75.30} & \underline{69.46} & \textbf{23.27} & \underline{5.19} & \underline{92.50} & \underline{46.17} & 0.00 & 35.30 & \textbf{22.95} & \textbf{14.30} & \textbf{39.28}           \\
        \midrule
        SwiftFuser$^{\dagger}$ & 80.65 & \underline{6.74} & 0.00 & 28.90 & \textbf{64.31} & \underline{91.15} & \underline{72.18} & 68.41 & \textit{13.12} & 1.28 & \textbf{92.86} & 42.32 & 0.00 & \textbf{51.81} & 19.33 & 7.89 & \underline{38.56}          \\
        
        {\color{blue}{
        RODSNet$^{\dagger}$+GRU
        }}
        & 80.82 & 4.09 & 0.00 & 30.67 & 57.55 & 90.82 & 68.09 & 68.70 & 16.57 & 1.96 & 93.23 & 42.75 & 0.00 & 58.86 & 21.97 & 7.99 & 38.73          
         \\
        
        \bottomrule
        \multicolumn{18}{r}{\footnotesize{$^{\dagger}$ResNet-18, $^{\ddagger}$ResNet-34 backbone}}
        \end{tabular}
        }
        \label{tab:semseg_results}
    \end{table*}
    
    \begin{table}[!t]
        \centering
        \caption{Real-time Validation and Test Semantic Segmentation Benchmark Results over all 17 classes and the obstacle class, in order of ascending test obstacle IoU.}
        \resizebox{\linewidth}{!}{%
        \begin{tabular}{rcccccc}
            
            
            

            \toprule
            \multirow{2}{*}{{\bf Network}} & \multicolumn{2}{c}{\bf{Val.}} & \multicolumn{2}{c}{\bf{Test}} & \multirow{2}{*}{{\bf \shortstack{Params\\(M)}}} &\multirow{2}{*}{{\bf \shortstack{Runtime\\(ms)}}}\\
            \cmidrule(lr){2-3} \cmidrule(lr){4-5}
             & mIoU & Obs. IoU & mIoU & Obs. IoU
            &  \\ 
            \midrule

            ESPNet~\cite{mehta2018espnet} 
            & 38.08 & 21.88 & 27.83 & 15.31 & 0.20 & 32  \\ 
            ENet~\cite{paszke2016enet}  
            & 43.73 & 37.93 & 33.30 & 15.56 & 0.35 & 86  \\ 
            MobileNetV3~\cite{howard2019searching} 
            &31.70 & 22.83 & 25.57 & 16.61 & 3.18 & 48 \\  
            LiteSeg$^{\dagger}$~\cite{emara2019liteseg} 
            & 41.17 & 37.15 & 29.02 & 17.60 & 3.51 & 69 \\
            CGNet~\cite{wu2020cgnet}  
            & 48.29 & 48.31 & 32.79 & 21.54 & 0.50 & 73 \\
            EDANet~\cite{lo2019efficient}  
            & 49.18 & 55.21 & 34.53 & 21.67 & 0.69 & 43  \\ %
            DenseNet$^{\star}$~\cite{krevso2020efficient}
            & 44.56 & 42.47 & 29.37 & 21.81 & 10.31 & 110 \\ 
            RPNet~\cite{chen2019residual}  
            & 43.93 & 30.19 & 32.37 & 25.08 & 1.89 & 60 \\ 
            FSFNet~\cite{kim2020accelerator} 
            & 44.06 & 45.20 & 33.73 & 26.24 & 0.83 & 43 \\ 
            LiteSeg$^{\ddagger}$~\cite{emara2019liteseg}  
            & 51.44 & 58.60 & 34.39 & 28.33 & 20.55 & 50   \\
            FC-HarDNet~\cite{chao2019hardnet}  
            & 47.29 & 53.88 & 32.60 & 28.51 & 4.12 & 9  \\ %
            FasterSeg~\cite{chen2019fasterseg}  
            & 44.21 & 43.41 & 28.70 & 29.44 & 5.67 & 46  \\ 
            ERFNet~\cite{romera2017erfnet}  
            & 49.15 & 55.76 & 37.02 & 30.58 & 2.07 & 53 \\  
            DeeplabV1~\cite{chen2014semantic}  
            & 49.23 & 57.90 & 33.44 & 33.19 & 20.50 & 30 \\
            LiteSeg$^{*}$~\cite{emara2019liteseg} 
            & 44.73 & 46.61 & 32.23 & 34.03 & 4.38 & 50  \\
            FCN~\cite{long2015fully}  & 60.71 & 38.59 & 38.59 & 34.94 & 18.64 & 22\\ 
            DeeplabV3~\cite{chen2017rethinking} 
            & 43.67 & 49.31 & 30.64 & 40.32 & 39.04 & 7 \\
            SegNet~\cite{badrinarayanan2017segnet}  
            & 57.91 & 62.85 & 36.49 & 40.60 & 29.45 & 5 \\ 
            DenseASPP~\cite{yang2018denseaspp} 
            & 50.57 & 60.96 & 34.77 & 42.05 & 10.22 & 12 \\
            PSPNet~\cite{zhao2017pyramid} 
            & 54.41 & 71.87 & 36.40 & 47.37 & 65.47 & 62 \\
            SwiftNet~\cite{orvsic2021efficient}  
            & 55.96 & 67.77 & 36.11 & 47.51 & 12.06 & 33 \\ 
            ESANet$^{\bigoplus}$~\cite{seichter2021efficient}
            & 65.41 & 87.85 & 41.53 & 48.80 & 23.94 & 10  \\ 
            BiSeNet~\cite{yu2018bisenet}
            & 53.15 & 64.64 & 35.34 & 51.23 & 12.56 & 6 \\
            \bottomrule
            \multicolumn{7}{r}{\footnotesize{$^{\dagger}$ShuffleNet~\cite{zhang2018shufflenet} $^{\ddagger}$DarkNet-19~\cite{darknet13} $^{*}$MobileNet V2~\cite{sandler2018mobilenetv2} $^{\star}$ Ladder-style} $^{\bigoplus}$RGB-D}
            
        \end{tabular}
        }
        \label{tab:real_time_semantic_segmentation}
    \end{table}

    \subsection{Waypoint Prediction}
        Waypoint prediction is defined as the task of navigating through the pre-defined set of routes. These route sequences contain sparse goal locations in GPS coordinates from the global planner, and the driving proficiency of an autonomous agent is evaluated based on these routes. Starting at an initial position, agents are to drive through the planned path to a destination point, where routes are derived from multiple domains, including freeways, urban scenes, and residential districts. Following the evaluation criteria from the Longest6 benchmark~\cite{Chitta2022PAMItransfuser}, we report the two multi-modal fusion transformer network performance on our dataset in Table~\ref{tab:waypoint_longest6_benchmark_results}.

        \noindent
        \textbf{Evaluation Metrics.} \quad
        \textit{Route completion} (RC) score is the percentage of route distances completed by the agent averaged across $N$ routes, with the penalty 1(-\% off route distance) given if the agent drives outside the planned route lanes for a percentage of the route (Eq.~\ref{eq:RC}).
        \begin{equation}
            \textbf{RC} = \frac{1}{N} \sum_{i}^{N} R_{i}.
            \label{eq:RC}
        \end{equation}
        
        During the route a penalty is given to the agent, starting with an ideal 1.0 base score, as a geometric series of infraction penalty coefficients ($p^{j}$) for every infraction instance $j$, \textit{i.e.,} \textit{Infraction Score} (IS) (Eq.~\ref{eq:IS}). The pre-defined penalty coefficients for collisions with a pedestrian, a vehicle and static layout, and for red light violations are 0.50, 0.60, 0.65, and 0.7, respectively.
        \begin{equation}
            \textbf{IS} = \prod_{j}^{\textbf{Ped,Veh,Stat,Red}} (p^{j})^{\text{infractions}^{j}}.
            \label{eq:IS}
        \end{equation}
        
        \textit{Driving Score} (DS) is the averaged route completion (RC) with infraction multiplier $P_{i}$ (Eq.~\ref{eq:DS}):
        
        \begin{equation}
            \textbf{DS} = \frac{1}{N} \sum_{i}^{N} R_{i}P_{i}.
            \label{eq:DS}
        \end{equation}
        
        \noindent
        \textbf{TransFuser} \quad
        is a recent network architecture for end-to-end driving consisting of multi-modal fusion transformer blocks across image and LiDAR representations, and an auto-regressive waypoint prediction network. Its success in road navigation using image and LiDAR data with the auxiliary task losses, such as those accounting for semantic segmentation and depth estimation, closely resembles and motivated our approach of refining the predicted image and depth representations. In their architecture, the respective representations from each modality are sequentially fused via a transformer (GPT) block, whose outputs are then added back to the respective input features. This stage is repeated four times and the final image and LiDAR representations are element-wise summed to feed into the auto-regressive waypoint prediction network. As a baseline comparison, the authors also introduce Latent TransFuser which replaces the 2-channel LiDAR histogram input (BEV) with an equal channel positional embedding of the same dimensions (256$\times$256$\times$2), whose values for the left-right and top-down axes each equally range from -1 to 1 in a single channel.

        For the waypoint prediction task on, we compared the TransFuser-based models along with our experimented models in Table~\ref{tab:waypoint_longest6_benchmark_results}. With image and LiDAR sensor data together in TransFuser, the results by the three metrics (DS, RC, IS) are higher by a large margin than those from LatentFuser. This insinuates the effectiveness of multi-modal feature fusion between image and LiDAR data using our dataset as well, in addition to their tested Town05 benchmark. Using our transformer-based fusion refinement, the route navigation scores are even higher. In particular, the higher route completion score indicates that SwiftFuser is capable of avoiding obstacles and arrive at the destination points more effectively. RODSNet+GRU is a based on refinement using a stacked hourglass structure followed by an auto-regressive waypoint prediction network with the fused representations as its input. We observed that a stacked hourglass structure may be more effective than the one-way transformer-based cross-modal learning in SwiftFuser. 
        
        \begin{table}[!h]
            \centering
            \caption{Route navigation benchmark test results evaluated on the test set of AVOID.}
            \begin{tabular}{lccc}
                \toprule
                {\bf Waypoint Prediction Model} & {\bf DS} $\uparrow$ & {\bf RC} $\uparrow$ & {\bf IS} $\uparrow$ \\ 
                \midrule
                
                Latent TransFuser~\cite{Chitta2022PAMItransfuser} & 79.40 & 86.41 & 0.88 \\
                TransFuser~\cite{Chitta2022PAMItransfuser} & 88.82 & 96.03 & 0.91 \\
                
                
                SwiftFuser & 94.22 & 98.07 & 0.95 \\
                
                RODSNet~\cite{song2022rodsnet}+GRU & 97.14 & 100.00 & 0.97 \\ 

                \midrule
                Expert (Oracle) & 97.86 & 100.00 & 0.98 \\
                \bottomrule
                
            \end{tabular}
        
            \label{tab:waypoint_longest6_benchmark_results}
        \end{table}
    
\section{Multi-task Learning on AVOID}
\label{sec:experiments}

\subsection{Overview}
We propose to examine the effect of transformer-based feature refinement against the stacked hourglass-style refinement as practiced in RODSNet~\cite{song2022rodsnet}. As in Fig.~\ref{fig:overview}, we employ the general structure of the current state-of-the-art obstacle detection network, RODSNet, in which disparity prediction is performed by a real-time disparity prediction network, AANet~\cite{xu2020aanet} yielding disparity maps of scales $\frac{1}{4}$, $\frac{1}{8}$, and $\frac{1}{16}$, given the multi-scale pyramidal features extracted sequentially over four encoder blocks. The refinement structure closely follows the Transfuser~\cite{Chitta2022PAMItransfuser}, whereby multiple sensor modalities are fused together in order to guide each other as contexts. In our experimented SwiftFuser network, we experimentally designed the network such that the predicted disparity maps are not to be fused with the outputs of each transformer block. We have conducted experiments on  (1) adding to each transformer block the respective disparity map of the same scale as the semantic map input, and (2) adding only the smallest disparity map of $\frac{1}{16}$ scale and propagating through the final prediction and to the input to the waypoint prediction network, and experimentally concluded that disparity maps interfere with semantic segmentation prediction severely. As shown in the intermediate-stage disparity maps in the figure, the quality is degraded over each fusion stage, indicating ineffectiveness of incorporating disparity into semantic segmentation prediction. 

\noindent
\textbf{Experimental Setup.} \quad
%
%
We conducted experiments in PyTorch 1.11, CUDA 11.3 environment. For the training loss, we used the semantic and disparity losses, altogether termed in $\mathcal{L}_{RODSNet}$ from \cite{song2022rodsnet}, with an $L_{1}$ loss for the waypoint $\mathbf{w}$ as described in Eq.~\ref{eq:loss}. The default learning rate (1e-4), step scheduler with weight decays, and the AdamW optimizer are conducted as in \cite{Chitta2022PAMItransfuser} for 40 training epochs.

\begin{equation}
    \mathcal{L} = \mathcal{L}_{\text{RODSNet}} + \sum_{t=1}^{T} ||\mathbf{w}_{t}-\mathbf{w}_{t}^{gt}||_{1}.
    \label{eq:loss}
\end{equation}





\begin{figure}[!h]
    \centering
    \includegraphics[width=\linewidth]{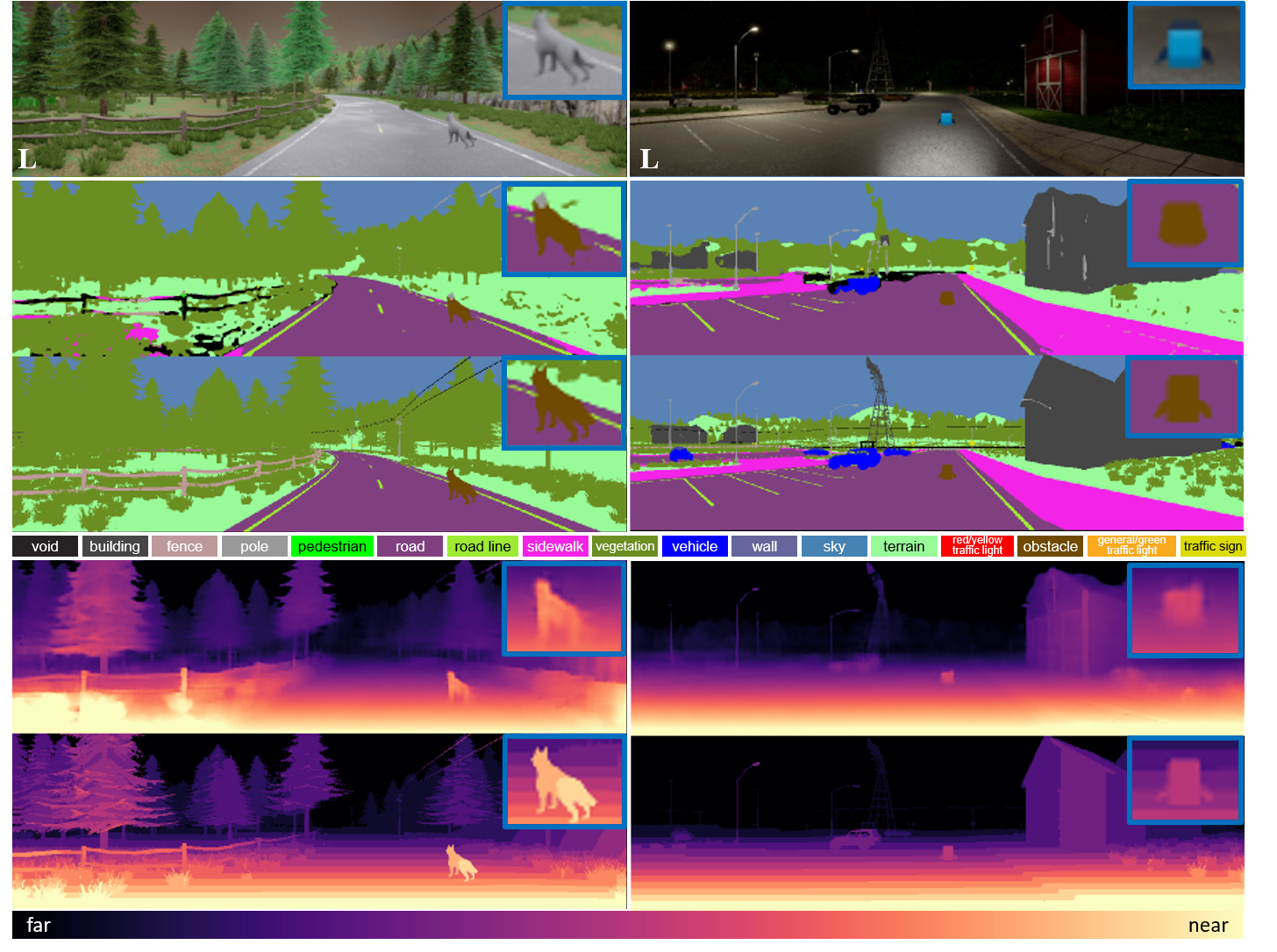}
    \caption{Obstacle detection results in test dataset using RODSNet+GRU. Given a stereo RGB pair input (only left is shown in \textit{top row}), the network detects unexpected obstacles from the predicted segmentation map (\textit{row 2}; GT in \textit{row 3}) as well as the predicted disparity map (\textit{row 4}; GT in \textit{bottom row})}
    \label{fig:unseen_obstacles}
\end{figure}

\noindent
\textbf{Evaluation Results.}
Using the ResNet-34 as the backbone upon fusing features, we observed that adding $\frac{1}{8}$ and $\frac{1}{4}$ scale features to the initial predicted segmentation map yields the highest obstacle IoU, with all other approaches running inference comparably. We posit that the lower performance with adding only $\frac{1}{16}$ in the early fusion stage deteriorates the image segmentation representations, while fusion at later stages exploits the disparity contexts in a global scope. We also qualitatively verify in Fig.~\ref{fig:unseen_obstacles} that RODSNet+GRU model is capable of distinguishing untrained obstacles from the test set, confirming the generalizability of the model. We report on depth range-wise obstacle detection performance in \textit{Supplementary}.

\section{Conclusion}
\label{sec:conclusion}

In this work, we introduced AVOID, a rich obstacle dataset for urban scene understanding with diverse, high-quality annotations to date. Our dataset encompasses the obstacle detection and road navigation tasks with obstacle avoidance in a simulated environment, providing semantic, geometric and LiDAR sensor data that are applicable in most visual perception tasks for autonomous driving. We hope our dataset serves as a new challenging benchmark for obstacle detection under adverse conditions and can mitigate the bottleneck problem due to the lack of large-scale obstacles datasets for robust visual perception. We also hope to extend our dataset for joint learning with other datasets collected in similar simulated domains.

{\small
\bibliographystyle{ieee_fullname}
\bibliography{references}
}

\end{document}